\documentclass[10pt,twocolumn,letterpaper]{article}

\usepackage{cvpr}              
\definecolor{cvprblue}{rgb}{0.21,0.49,0.74}

\usepackage{url}
\definecolor{gray}{rgb}{0.92, 0.92, 0.92}
\definecolor{yellow}{rgb}{1, 1, 0.7}
\definecolor{orange}{rgb}{1, 0.85, 0.7}
\definecolor{red}{rgb}{1, 0., 0.}

\usepackage{multirow}
\usepackage{graphicx}
\usepackage{booktabs}
\usepackage{placeins} 
\usepackage{siunitx}
\usepackage{enumitem}
\usepackage{subcaption}

\usepackage[pagebackref,breaklinks,colorlinks,allcolors=cvprblue]{hyperref}

\def\paperID{5749} 
\def\confName{CVPR}
\def\confYear{2026}

\title{GenHOI: Towards Object-Consistent Hand–Object Interaction with Temporally Balanced and Spatially Selective Object Injection}

\author{Xuan Huang\textsuperscript{1,2}\footnotemark[1], Xiang Mochu\textsuperscript{1,3\footnotemark[1]}, Zhelun Shen\textsuperscript{1}\footnotemark[2], Jinbo Wu\textsuperscript{1}, Chenming Wu\textsuperscript{1} \\
Chen Zhao\textsuperscript{1}, Kaisiyuan Wang\textsuperscript{1}, Hang Zhou\textsuperscript{1}, Shanshan Liu\textsuperscript{1}, Haocheng Feng\textsuperscript{1}, Wei He\textsuperscript{1}, Jingdong Wang\textsuperscript{1}\\
\textsuperscript{1}Department of Computer Vision Technology(VIS), Baidu Inc., China \\ \textsuperscript{2}Shenzhen Campus of Sun Yat-Sen University, China 
\textsuperscript{3}Northwestern Polytechnical University, China
\\
{\tt\small shenzhelun@pku.edu.cn}
}
\begin{document}
\twocolumn[{
\maketitle
\begin{center}
  \vspace{-0.2in}
  \includegraphics[width=0.9\textwidth]{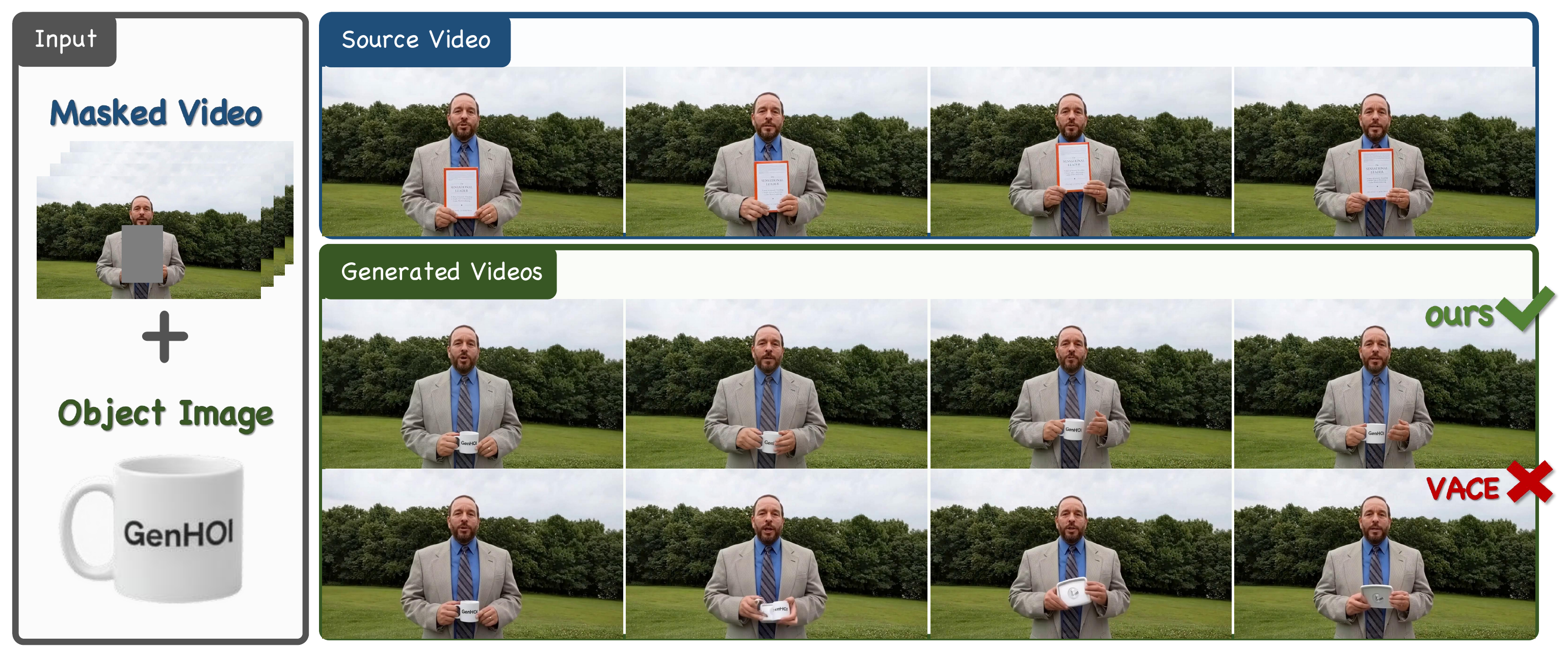}
  \vspace{-0.1in}
  \captionof{figure}{
  \textbf{Comparison with representative method.} 
  All-in-one video editing models (\textit{e.g.}, VACE) benefit from large-scale Internet training data, they still struggle to maintain object consistency across frames. 
  In contrast, our method achieves both strong generalization and natural, visually consistent interactions between the human and the object. 
  }
  \label{fig:banner}
\end{center}
}]

\begin{abstract}
Hand–Object Interaction (HOI) remains a core challenge in digital human video synthesis, where models must generate physically plausible contact and preserve object identity across frames. Although recent HOI reenactment approaches have achieved progress, they are typically trained and evaluated in-domain and fail to generalize to complex, in-the-wild scenarios. In contrast, all-in-one video editing models exhibit broader robustness but still struggle with HOI-specific issues such as inconsistent object appearance. In this paper, we present GenHOI, a lightweight augmentation to pretrained video generation models that injects reference-object information in a temporally balanced and spatially selective manner. 
For temporal balancing, we propose Head-Sliding RoPE, which assigns head-specific temporal offsets to reference tokens, distributing their influence evenly across frames and mitigating the temporal decay of 3D RoPE to improve long-range object consistency. For spatial selectivity, we design a two-level spatial attention gate that concentrates object-conditioned attention on HOI regions and adaptively scales its strength, preserving background realism while enhancing interaction fidelity. Extensive qualitative and quantitative evaluations on unseen, in-the-wild scenes demonstrate that GenHOI significantly outperforms state-of-the-art HOI reenactment and all-in-one video editing competitors. Project Page: \url{https://xuanhuang0.github.io/GenHOI/}.
\end{abstract}

\makeatletter
\renewcommand{\thefootnote}{\fnsymbol{footnote}}
\makeatother
\footnotetext[1]{Equal contribution. Work was done when intern at Baidu.}
\footnotetext[2]{Corresponding authors. Zhelun Shen is the project lead.}
\section{Introduction}

Hand–Object Interaction (HOI) is a fundamental aspect of digital human content creation, particularly in domains such as online education and e-commerce. Although significant progress has been made in digital human generation, achieving realistic HOI remains a major challenge due to the high variability in object shapes, human poses, and interaction orientations.  
Consider a scenario in which a designer aims to combine a product image into a real-world broadcast video, as illustrated in Fig.~\ref{fig:banner}. Two objectives must be met to achieve a convincing result. First, the interaction between the hand and the object should appear natural and physically plausible. Second, the visual consistency of the reference object, such as color, texture, and logo, must be preserved across frames. Realizing these objectives is difficult because the shapes, poses, and orientations of objects vary widely, requiring the model to reason about physical contact and contextual semantics.

Recent HOI reenactment works have made encouraging progress. For example, HOI-Swap~\cite{hoi-swap} extends image-level inpainting to the video domain by sequential frame warping, while Re-HOLD~\cite{rehold} introduces a layout-guided diffusion model for controllable HOI synthesis. However, these methods are typically trained and evaluated on in-domain data and fail to generalize to real-world in-the-wild scenes.  
All-in-one video editing models, such as VACE \cite{jiang2025vace} exhibit stronger generalization due to their large-scale internet data pretraining. Nevertheless, they still struggle with HOI-specific challenges, e.g., maintaining object consistency. These limitations highlight the need for a model that generalizes robustly to unseen environments while ensuring both realism and identity preservation.

In this paper, we propose GenHOI to address the aforementioned challenges. GenHOI aims to improve HOI reenactment beyond source-domain datasets and extend it to challenging in-the-wild scenarios. To this end, we augment existing video generation models with a lightweight module designed for temporally balanced and spatially selective object information injection. This approach enables the network to learn the HOI task more effectively while minimizing degradation to its original video generation capabilities. For temporal balancing, we introduce \textit{Head-Sliding RoPE}, which assigns head-wise temporal offsets to reference-object tokens. This method distributes their influence evenly across frames, thereby mitigating temporal degradation of the reference response when injected into video tokens during self-attention. For spatial selectivity, we introduce a spatial attention gate that controls where information flows and how strongly it is applied across the spatial dimension.

We validate our approach through comprehensive qualitative and quantitative evaluations, all performed on unseen scenes to ensure a fair assessment of generalization. The results show that our framework significantly outperforms state-of-the-art HOI reenactment methods and all-in-one video editing models, achieving higher realism and smoother hand motions and interactions. Figure~\ref{fig:banner} presents a visual comparison with our strongest competitor, VACE\cite{jiang2025vace}, confirming our clear advantage in object consistency. This improvement primarily stems from our design, which injects reference-object information in a temporally balanced and spatially selective manner.
We summarize the contributions of this paper as follows.
\begin{itemize}
\item We introduce Head-Sliding RoPE, which assigns head-specific temporal offsets to reference tokens to balance their influence across frames, mitigating the temporal decay of 3D RoPE and improving object consistency in long video generation.
\item We design a two-level spatial attention gate that concentrates object-conditioned attention on HOI regions and adaptively scales its strength, improving interaction fidelity while preserving background realism.
\item Through the synergy of Head-Sliding RoPE and the spatial attention gate, we augment existing video generation models with a lightweight module that provides temporally balanced, spatially selective object information injection. This boosts hand–object realism and object-identity consistency while minimizing degradation to the base model’s video generation capabilities.
\end{itemize}

\section{Related Work}

\begin{figure*}[t]
\includegraphics[width=1.0\linewidth]{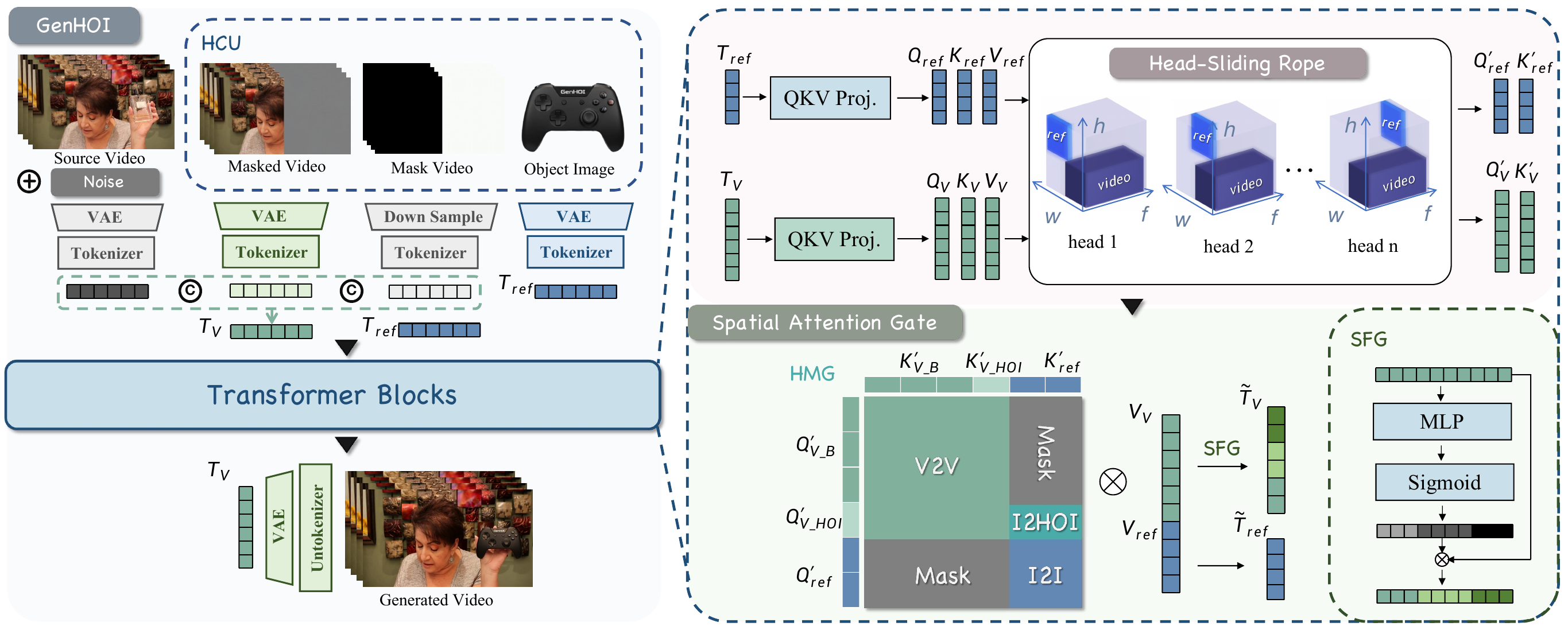}
\vspace{-0.1in}
\caption{
\textbf{Overview of the proposed framework.} The model integrates the HOI Condition Unit, Head-Sliding RoPE, and Spatial Attention Gate for temporally balanced and spatially selective HOI reenactment. HMG denotes hard mask gate and SFG is the soft flow gate.}
\vspace{-0.1in}
\label{fig:pipeline}
\end{figure*}

\subsection{Human Body Animation}
The rapid advancement of diffusion models has greatly accelerated progress in human body animation. A major line of research builds upon U-Net architectures~\cite{unet} enhanced with cross-attention mechanisms~\cite{attention} to integrate contextual or textural information into the generation process. Among early works, DreamPose~\cite{dreampose} employs UV maps as motion signals and conditional embeddings for motion transfer, inspiring subsequent methods such as AnimateAnyone~\cite{animateanyone}, DISCO~\cite{disco}, MimicMotion~\cite{zhang2024mimicmotion}, MagicPose~\cite{magicpose}, ShowMaker~\cite{yang2025showmaker}, and TalkAct~\cite{talkact}.
Recently, diffusion transformers (DiT) have demonstrated superior scalability, leading to DiT-based architectures for human animation, including HumanDiT~\cite{gan2025humandit}, Human4DiT~\cite{shao2024human4dit}, UniAnimate~\cite{wang2025unianimate}, and DreamActor-M1~\cite{luo2025dreamactor}. While human–object interaction (HOI) reenactment can be viewed as a specialized subtask of human animation, this general line of work still faces challenges in maintaining consistent human identity and object appearance across long video sequences. This highlights the need for models explicitly designed for the unique constraints of HOI reenactment.

\subsection{Diffusion-based Video Editing and Inpainting}
A substantial body of research has focused on adapting pre-trained diffusion models for video editing and inpainting. Many frameworks~\cite{stablediffusion, tokenflow, emu, rerender, fatezero, pix2video, space, tuneavideo, shape, ccedit, editavideo, videop2p, makeavideo, videogeneration} rely on textual prompts for control. While convenient, this approach often lacks the precision needed to describe fine-grained object details or spatial relationships especially in complex Human–Object Interaction (HOI) scenarios. Recently, VideoPainter~\cite{bian2025videopainter} further advances the field with a DiT-based dual-branch architecture that separates foreground generation from background preservation.
Despite these advances, most methods still depend on text inputs, which can be ambiguous. Concurrently, VACE~\cite{jiang2025vace} was the first to introduce an all-in-one model for video creation and editing that supports diverse condition combinations (including image reference and inpainting), a paradigm similar to ours. Despite their powerful, general-purpose capabilities, these 'all-in-one' models still struggle with HOI-specific challenges, particularly maintaining object consistency and fidelity.


\subsection{Human-Object Interaction}
Alongside recent advances in broad human-centric tasks like head-body animation and lip-syncing~\cite{wav2lip,lsp,vprq,liveportrait,pcavs,LIA,facevid2vid,vpgc,emo,hallo}, research is also beginning to address the more fine-grained challenge of modeling HOI~\cite{hoi, affordance, diffusionhoi, hoidiffusion, cghoi, graspxl, lego}. These interactions are critical for achieving realistic human behavior in both static and dynamic settings.
Among these, DiffHOI~\cite{diffusionhoi} employs a diffusion-based framework to model the conditional distribution of object renderings, thereby facilitating novel-view synthesis of HOIs. GraspXL~\cite{graspxl} offers a unified framework for generating hand-object grasping motions across a range of object geometries and hand morphologies. Cg-HOI~\cite{cghoi} focuses on generating plausible 3D human-object interactions from textual descriptions, enhancing controllability in semantic-to-motion generation. HOI-Swap~\cite{hoi-swap} extends static image-level HOI inpainting to a video-based framework by incorporating sequential frame warping. Although these models achieve impressive performance on in-domain datasets, they often struggle to generalize to in-the-wild scenarios. This limitation largely stems from the heavy reliance on task-specific priors and extensive modifications to the base foundation models.

\section{Proposed Method}

\subsection{Overview}
\label{sec:method:overview}
Given a source video $\mathbf{V}=\{I_1,\ldots,I_F\}\in\mathbb{R}^{F\times H\times W\times 3}$ and an object reference image $I_{\mathrm{ref}}\in\mathbb{R}^{H\times W\times 3}$, our goal is to reenact realistic hand--object interactions (HOI) between hands and the target object. 
Our framework is trained in a self-supervised reconstruction manner, where the reference video \( \mathbf{V}_\text{r} \in \mathbb{R}^{F \times H \times W \times 3} \), the binary mask video \(\mathbf{V}_\text{mask}\) and the reference image  \( I_{\text{ref}} \) are derived from the original source video \( \mathbf{V} \).
The training objective is to reconstruct \( \mathbf{V} \) from random noise $\mathbf{X}_{\mathrm{rand}}$ conditioned on {\( V_{r} \)}, \( I_{\text{ref}} \) and \(\mathbf{V}_\text{mask}\), formulated as: 
\begin{equation}
\mathbf{V} = \mathcal{D}(\mathcal{M}(\mathbf{X}_\text{rand}, \mathcal{E}(I_\text{ref}), \mathcal{E}(V_{r}), \psi(\mathbf{V}_{\text{mask}}))),
\end{equation}
We employ a pretrained VAE encoder–decoder $(\mathcal{E},\mathcal{D})$ to compress videos into latent representations and reconstruct outputs.
while a DiT-based denoiser $\mathcal{M}$ predicts iterative denoising. The operator $\psi$ denotes an average pooling operation.  

To adapt a general video generation model to the HOI reenactment task, we introduce a HOI Condition Unit that injects HOI-related cues in a manner consistent with the original input organization. Next, built upon the DiT framework, we propose Head-Sliding RoPE and a spatial attention gate to inject reference object information in a temporally balanced and spatially selective manner. During inference, a novel object reference image is supplied to reenact the target video. An overview of the pipeline is shown in Fig.~\ref{fig:pipeline}. We describe the HOI Condition Unit in Sec.~\ref{sec:Unified Inpainting-based latent Process} and the temporally balanced, spatially selective attention mechanism in Sec.~\ref{sec:HOI-centric Attention Mechanism}.

\subsection{HOI Condition Unit}
\label{sec:Unified Inpainting-based latent Process}

To adapt a pretrained video generation model to the HOI reenactment task without introducing additional network branches or parameters, we propose a HOI Condition Unit (HCU). This module injects HOI-relevant cues directly into the latent input stream, preserving the model’s original generative capacity while enabling effective HOI-specific learning.

Given an input video \(\mathbf{V}\) and its corresponding binary mask video \(\mathbf{V}_\text{mask}\), we first construct the reference video $\mathbf{V}_\text{r}$ that defines the inpainting regions:
\begin{equation}
    \mathbf{V}_\text{r} = 
    \begin{cases}
        \mathbf{V}, & F = 0, \\
        (1 - \mathbf{V}_\text{mask}) \cdot \mathbf{V} + \mathbf{V}_\text{mask} \cdot \lambda, & F > 0,
    \end{cases}
\end{equation}
where $\mathbf{V}_\text{mask}$ identifies the hand–object interaction (HOI) regions, $\lambda$ is a constant, set to 127 in our network, which becomes 0 after normalization, and $F$ denotes the frame index in the input video. We then project all inputs into the latent space and concatenate them along the channel dimension to form the input of the DiT model:
\begin{equation}
\mathbf{L_v} = \text{Concat}\!\left(\mathbf{X_t},\, \mathcal{E}(\mathbf{V}_{\text{r}}),\, \psi(\mathbf{V}_{\text{mask}}) \right),
\end{equation}
where $\mathbf{X_t}$ represents the noisy target video latent.
In this way, we reformulate the video generation process as a \textit{first-frame-conditioned video inpainting task}, where reference video $\mathcal{E}(\mathbf{V}_\text{r})$ provides background context and mask video $\psi(\mathbf{V}_\text{mask})$ tells the model where edits should occur. This formulation enables the model to better preserve background consistency while focusing generation on the HOI regions.

\begin{figure}[!t]
 \centering
 \includegraphics[width=1\linewidth]{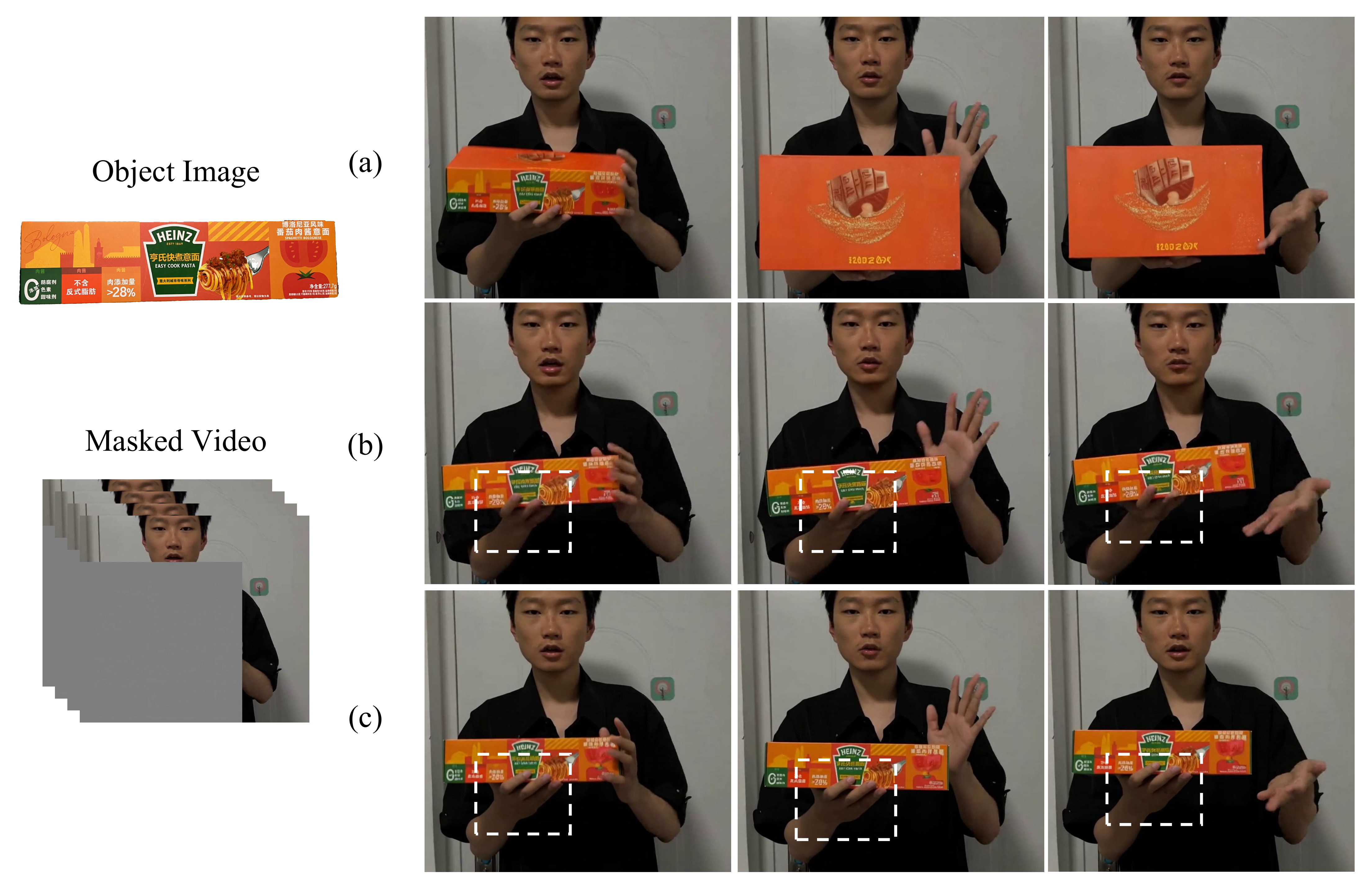}
 \vspace{-0.2in}
 \caption{\footnotesize Visualization comparison between different reference object injection methods: (a) only HOI condition unit (HCU); (b) HCU + ref-in-bbox conditioning; (c) HCU + the proposed Attention.}
 \label{fig:ablation_ref}
  \vspace{-0.1in}
\end{figure}

\subsection{Temporally Balanced, Spatially Selective Attention}
\label{sec:HOI-centric Attention Mechanism}
A key challenge of HOI is how to effectively inject reference object information into the model to ensure visual consistency across diverse object shapes and orientations. A straightforward approach is to incorporate this information from the first frame. However, as shown in Fig.~\ref{fig:ablation_ref}(a), relying solely on the first frame often leads to noticeable object degradation over time.

Another commonly adopted strategy is the “ref-in-bbox” method, which directly pastes the reference object image into the inpainting region of $\mathbf{V}_\text{r}$:
\vspace{-0.03in}
\begin{equation} 
\mathbf{V}_\text{r} = \begin{cases} \mathbf{V}, & F = 0, \\ (1 - \mathbf{V}_\text{mask}) \cdot \mathbf{V} + \mathbf{V}_\text{mask} \cdot \mathbf{I}_\text{ref}, & F > 0, \end{cases}
\label{eq:ref_in_bbox}
\end{equation}
where $\mathbf{I}_\text{ref}$ denotes the object reference image.
As illustrated in Fig.~\ref{fig:ablation_ref} (b), the “ref-in-bbox” approach can improve object appearance consistency in long video sequences. However, this encourages the model to treat the task as simple inpainting rather than interaction. Consequently, the model fails to generate natural interaction, instead just ``pasting" the reference into the inpainting region.

In contrast, we propose Temporally Balanced, Spatially Selective Attention to address these issues. The overall pipeline is shown in the right part of Fig.~\ref{fig:pipeline}. Specifically, \textit{Head-Sliding RoPE} assigns head-wise temporal offsets to the reference object tokens, distributing their influence evenly across frames and mitigating temporal degradation of the reference response.  Meanwhile, a \textit{Spatial Attention Gate} focuses attention on interaction regions (e.g., HOI regions) while suppressing background interference.

\subsubsection{Head-Sliding RoPE} 

Rotary Positional Embedding (RoPE) \cite{su2024roformer} is widely adopted in DiT architectures to encode spatiotemporal positional information. 
We extend the standard video diffusion 3D RoPE to construct our Head-Sliding RoPE, which is applied to both query and key embeddings within self-attention layers. 
For the query embeddings, our RoPE is defined as
\begin{equation}
    \begin{bmatrix}
    Q_{re f}^{'} \\
    Q_{V}^{'}
    \end{bmatrix}  = 
    \begin{bmatrix}
    Q_{ref} \\
    Q_{V}
    \end{bmatrix} \odot [R_{f}, R_{h}, R_{w}]
\end{equation}
where $Q_{\text{ref}}\in\mathbb{R}^{N\times D}$ and $Q_{V}\in\mathbb{R}^{M\times D}$ denote the query embeddings for reference-image tokens and video tokens, respectively. The rotary embedding components $R_{f}\in\mathbb{R}^{(N+M)\times D_{1}}$, $R_{h}\in\mathbb{R}^{(N+M)\times D_{2}}$, and $R_{w}\in\mathbb{R}^{(N+M)\times D_{3}}$ encode the frame, height, and width dimensions.

In settings involving both video tokens and conditional (reference object) tokens, prior works \cite{xue2025stand, ominicontrol} typically assigns conditional tokens to a separate coordinate space, following the formulation of \cite{su2024roformer}, where $Q' = Q e^{j m \theta}$. Under this design, the formulation is as follows:
\begin{equation}
\scalebox{0.63}{
$
[R_{f}, R_{h}, R_{w}] =
\begin{bmatrix}
 \begin{pmatrix}
        e^{j (-1) \theta} \\
        \cdots \\
        e^{j (-1) \theta} \\
    \end{pmatrix}_{\scriptscriptstyle N \times D_{1}} \!\!\!\!\!\!\!\!,
    &
    \begin{pmatrix}
        e^{j (h^{0}_{I} + H_{V})\theta} \\
        \cdots \\
        e^{j (h^{N}_{I} + H_{V})\theta}
    \end{pmatrix}_{\scriptscriptstyle N \times D_{2}} \!\!\!\!\!\!\!\!,
    &
    \begin{pmatrix}
        e^{j (w^{0}_{I} + W_{V})\theta} \\
        \cdots \\
        e^{j (w^{N}_{I} + W_{V})\theta}
    \end{pmatrix}_{\scriptscriptstyle N \times D_{3}} 
    \\
     \begin{pmatrix}
        e^{j f^{0}\theta} \\
        \cdots \\
        e^{j f^{M}\theta}
    \end{pmatrix}_{\scriptscriptstyle M \times D_{1}}
    &
    \begin{pmatrix}
        e^{j h^{0}_{V}\theta} \\
        \cdots \\
        e^{j h^{M}_{V}\theta}
    \end{pmatrix}_{\scriptscriptstyle M \times D_{2}} \!\!\!\!\!\!\!\!,
    &
    \begin{pmatrix}
        e^{j w^{0}_{V}\theta} \\
        \cdots \\
        e^{j w^{M}_{V}\theta}
    \end{pmatrix}_{\scriptscriptstyle M \times D_{3}} 
\end{bmatrix}
$
}
\end{equation}





while $h_{I}^{i}$ and $w_{I}^{i}$ denote the height and width indices of the $i$-th reference image token, and $h_{V}^{i}$ and $w_{V}^{i}$ denote the indice of the video tokens. Here, $H_{V}$ and $W_{V}$ represent the spatial lengths of video tokens in their respective dimensions. Although this design enables the network to distinguish between video tokens and object reference tokens, it also has an inherent limitation. 
Due to the nature of RoPE, larger positional distances result in weaker attention responses. Consequently, when conditional tokens are assigned a fixed frame index (e.g., $-1$), their influence on video tokens becomes uneven across the temporal dimension. The response is strongest at the earliest frames and weakest at the latest ones. This imbalance leads to degradation of object fidelity in HOI reenactment.

To overcome this issue, we revise the formulation into the proposed Head-Sliding RoPE:


    

\begin{equation}
\scalebox{1.0}{
$
[R_{f}] =
\begin{bmatrix}
    \begin{pmatrix}
        e^{j\!\left(\left\lceil \frac{N_{f}}{N_{head}} n_{head} \right\rceil \right)\!\theta} \\
    \end{pmatrix}_{N \times D_{1}} \\
    \begin{pmatrix}
        e^{j f^{0}\theta} \\
        \vdots \\
        e^{j f^{m}\theta}
    \end{pmatrix}_{M \times D_{1}} \\
\end{bmatrix}
$
}
\end{equation}

where $n_{head}$ denotes the n-th head in the multi-head attention mechanism, $N_f$ is the total number of frames in the latent video representation, and $N_{head}$ is the total number of attention heads, $f^{i}$ is frame index.  
In this formulation, the frame index assigned to conditional tokens \emph{slides} (Visualization in Fig.\ref{fig:pipeline} Head-sliding RoPE) across different attention heads, effectively averaging their attention responses over the entire temporal span of the video. This design balances the interaction between conditional and video tokens across frames while maintaining distinct spatial coordinates, thereby enabling the model to differentiate token types and preserve high-fidelity object appearance throughout the sequence.


\subsubsection{Spatial Attention Gate}
We formulate HOI reenactment as a video inpainting problem, where the main challenge is to concentrate modeling capacity on the hand--object interaction (HOI) regions rather than the comparatively easy background. To this end, we introduce a two-level spatial attention gate.

\noindent \textbf{Hard Mask Gate}
Reference-object tokens are helpful for synthesizing hand–object interaction (HOI) regions but can be detrimental to background areas: within HOI regions, the model needs guidance from the reference tokens to determine which object to inject, whereas in background regions the reference video $\mathbf{V}_\text{r}$ already provides sufficient context; excessive influence from the reference tokens may introduce artifacts.

\begin{figure}[!t]
 \centering
 \includegraphics[width=1\linewidth]{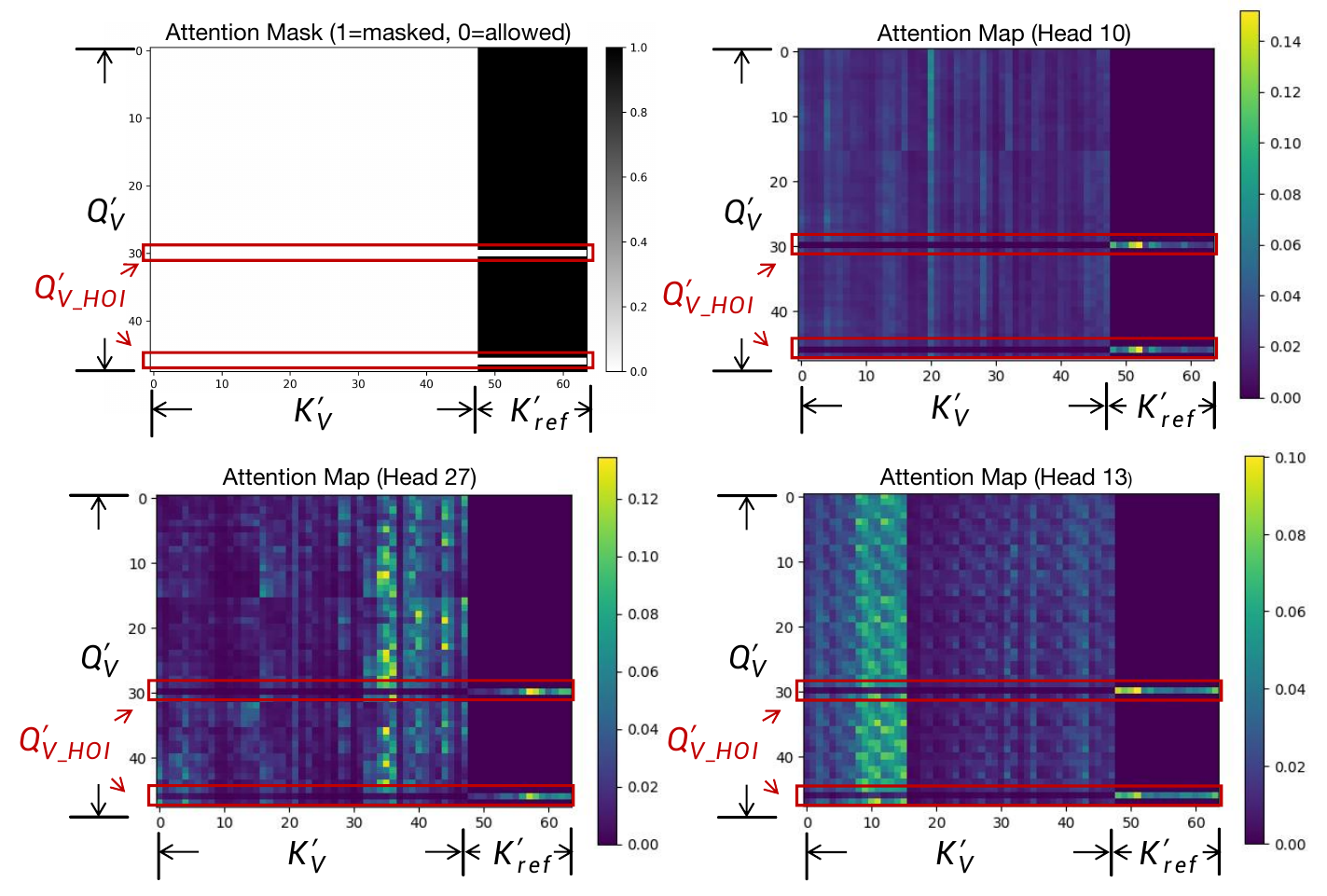}
 \vspace{-0.2in}
 \caption{Visualization of the applied hard mask gate and the resulting attention maps. Multiple attention maps across different heads are shown, highlighting the pronounced effect of the proposed mechanism. The red box indicates the interaction between queries in HOI regions and keys from both video and reference object tokens.}
 \label{fig:attn_mask}
\end{figure}

To address this, we apply a hard mask gate that enforces information flow only passes from reference-object tokens to HOI tokens while preventing their influence on background tokens.
Specifically, we denote the HOI-region and background-region video tokens as $T_{V_{HOI}}$ and $T_{V_{B}}$, respectively. For the mapped query, key, and value embeddings
$Q = [Q_{V_{HOI}}, Q_{V_{B}}, Q_{ref}]$,
$K = [K_{V_{HOI}}, K_{V_{B}}, K_{ref}]$,
$V = [V_{V_{HOI}}, V_{V_{B}}, V_{ref}]$,
where the subscripts $V_{HOI}$, $V_{B}$, and $ref$ denote HOI, background, and object tokens, respectively,
we introduce a binary mask $M$ to control the information flow during attention computation:
\begin{equation}
M_{m,n} = \left\{
\begin{array}{c l}
0 & \text{if} \ m\in \{Q_{V_{B}}\} \ \text{and} \ n \in \{K_{ref}\},  \\
0 & \text{if} \ m\in \{Q_{ref}\} \ \text{and} \ n \in \{K_{V_{HOI}}, K_{V_{B}}\}, \\
1 & \text{elsewhere.} \\
\end{array}
    \right.
\end{equation}
Here, ${m}$ and ${n}$ index the query and key tokens, respectively. The final attention output is given by
\begin{equation}
    T_{out} = \text{softmax} \left( \frac {M \odot Q K^\intercal} {\sqrt{d_k}} \right) V .
\end{equation} 
This hard mask gate permits HOI queries to attend to object keys while blocking background queries from doing so, and it prevents object queries from attending back to video keys. Consequently, information from reference tokens flows into HOI regions (or remains self-regularized) without contaminating the background. See visualization in HMG (Hard Mask Gate) part of Fig.\ref{fig:pipeline}.
Fig.~\ref{fig:attn_mask} visualizes the attention map $\operatorname{softmax}\left(\frac{M \odot (QK^{\top})}{\sqrt{d_k}}\right)$ after applying the hard mask gate. As shown, object keys yield zero responses to background queries, while producing strong activations for HOI-region queries, which are often even stronger than those induced by video keys. This observation confirms that the proposed hard mask gate effectively blocks information flow from the reference object to background regions and encourages HOI queries to rely on reference object keys.

\noindent \textbf{Soft Flow Gate.}
Beyond the hard truncation above, we add a soft, token-wise gate that dynamically modulates the updated video tokens. This design is inspired by the common practice in LLMs \cite{gate}. Specifically, we decompose the output as $T_{\text{out}} = [\,T'_{v},\, T'_{ref}\,]$, where $T'_{v}$ collects the updated video tokens. We then compute per-token gating coefficients via LayerNorm layer $\mathcal{LN}$ and FCN $\mathcal{F}$, followed by a sigmoid:
\begin{equation}
\begin{aligned}
G_{v} &= \sigma\!\big(\mathcal{F}(\mathcal{LN}(T'_{v}))\big),\\
\tilde{T}_{v} &= G_{v} \odot T'_{v},
\end{aligned}
\end{equation}
where $\sigma(\cdot)$ is the sigmoid and $\odot$ denotes element-wise multiplication. The gated representation $\tilde{T}_{v}$ adaptively amplifies informative regions and suppresses redundant responses. Corresponding visualization is shown in SFG (Soft Flow Gate) part in Fig.\ref{fig:pipeline}. 

In conclusion, the \textit{Hard Mask Gate} controls \emph{where} information can flow, and the \textit{Soft Flow Gate} scales \emph{how strong} it should be based on content. Their complementarity defines the \textit{Spatial Attention Gate}, which concentrates attention on HOI regions, protects the background, and injects reference cues robustly for high-fidelity reenactment.

\

\section{Experimental Results}


\subsection{Implementation Details}
\noindent\textbf{Implementation.}
We implement the proposed method on top of the pretrained Wan-14B-I2V model~\cite{wan2025wanopenadvancedlargescale}, which is denoted as \textit{Ours} in the following. Training is conducted on a dataset containing approximately 19,000 video using 16× NVIDIA H100 (80 GB) GPUs for three days. The learning rate is set to 1 $\times$ $10^{-5}$. During training, the reference object image and the first frame are extracted from the original video, while during inference, the reference image is provided by the user and the first frame is generated using image editing methods \cite{NanoBanana2025, flux_kontext}.



\vspace{0.5em}
\noindent \textbf{Evaluation Dataset.} 
We evaluate the performance of our approach on in-the-wild scenes using the AnchorCrafter dataset~\cite{xu2024AnchorCrafter}, which provides diverse human reenactment videos captured across various scenes and identities. To ensure high visual quality, we select videos with a resolution of at least 720p and reorganize the dataset to include 50 videos for self-reenactment and 50 videos for cross-reenactment. The resulting subset is referred to as AnchorCrafter\_HOI. 

\vspace{0.5em}
\noindent \textbf{Baselines.} 
We conduct a comprehensive comparison between our method and several state-of-the-art (SOTA) approaches. The compared methods cover different lines of work, including the human body animation models UniAnimate-DiT~\cite{wang2025unianimate} and MimicMotion~\cite{zhang2024mimicmotion}, the all-in-one video editing framework VACE~\cite{jiang2025vace}, and the HOI reenactment model HOI-Swap~\cite{hoi-swap}. In the cross-reenactment setting, we ensure a fair comparison by supplying the same edited first frame to all methods that require an initial frame.

\subsection{Evaluation Setting}

\begin{figure*}[t]
  \centering
  \includegraphics[width=0.9\linewidth]{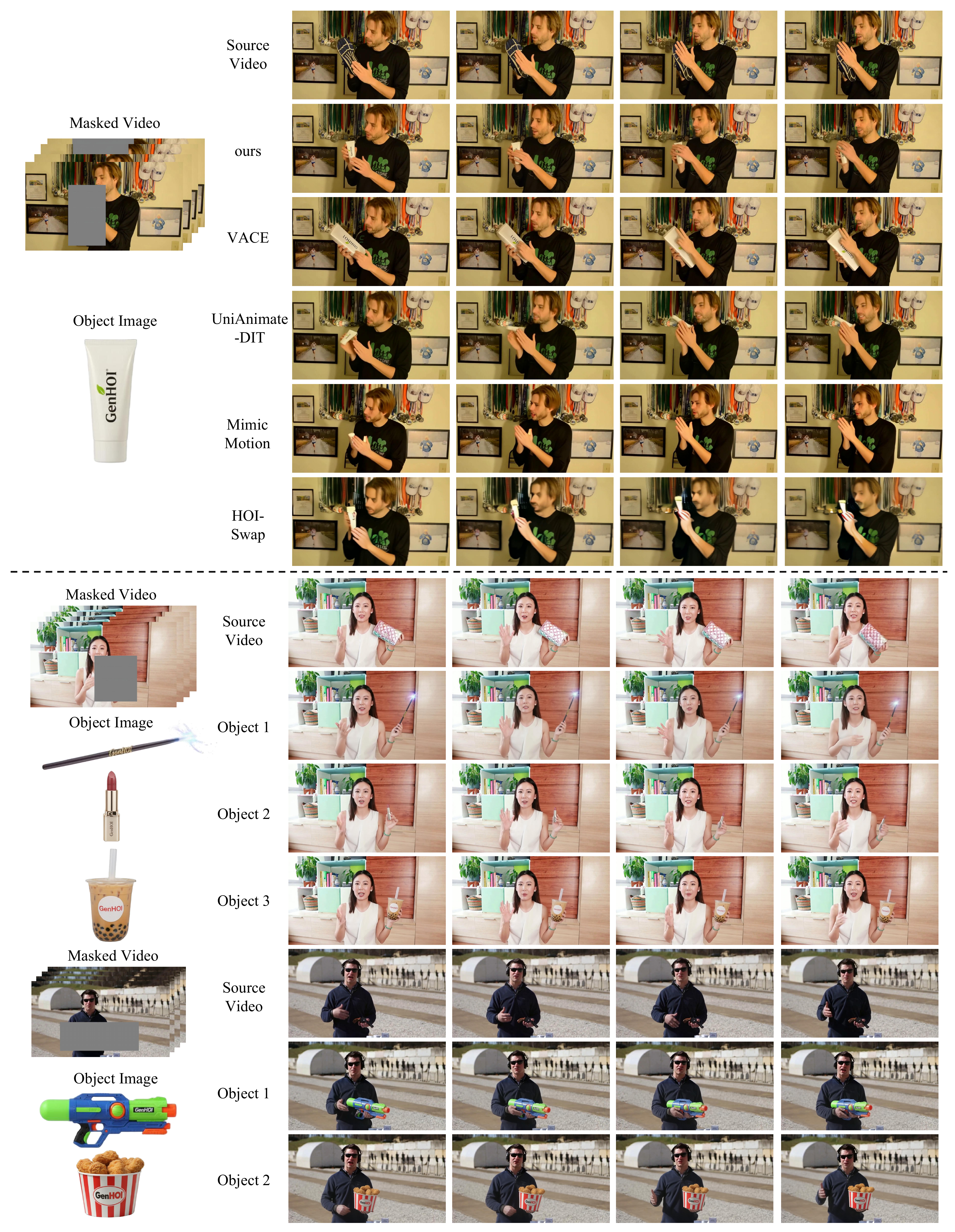}
  \caption{Up: Qualitative comparison with state-of-the-art methods. Down: Cross-reenactment results of the proposed method on in-the-wild videos, demonstrating robust and flexible object reenactment across various shapes, sizes, and categories.}
  \label{fig:swap}
\end{figure*}

We evaluate the proposed method in three settings:

\noindent\textbf{Self-reenactment.}
In this setting, the model reconstructs the original video from a masked source video and the corresponding object image of the video. We assess visual fidelity using PSNR~\cite{sara2019image}, SSIM~\cite{wang2004image}, and LPIPS~\cite{zhang2018unreasonable}; perceptual realism with FID~\cite{heusel2017gans} and FVD~\cite{unterthiner2018towards}; and object consistency via Object~CLIP (OC)~\cite{huang2025hunyuanvideo}, which computes CLIP similarity~\cite{radford2021learning} within the object bounding box.

\noindent\textbf{Cross-reenactment.}
In this setting, the method replaces the object in the source video with a novel one absent from the original sequence. Visual quality is evaluated using FID~\cite{heusel2017gans} and FVD~\cite{unterthiner2018towards}, and a user study is conducted due to the lack of ground truth.

\noindent\textbf{User Study.}
We conducted a user study with 30 participants who evaluated 40 short videos and 20 long videos from AnchorCrafter\_HOI. The participants rated the generated videos based on two criteria: (1) Reference fidelity with the reference image, (2) Video quality. Each criterion was assessed on a scale from 1 to 5, where 1 represents the lowest quality and 5 the highest. The user study was approved by the IRB.

\begin{table*}[!t]
  \centering
  \caption{Quantitative comparison on the AnchorCrafter\_HOI dataset. We evaluate performance under two settings: short video generation (81 frames) and long video generation (401 frames). VQ denotes video quality and RF denotes Reference Fidelity.}
  \label{tab:short_video_main_result}
  \resizebox{0.9\textwidth}{!}{%
    \begin{tabular}{clcccccc|cc|cc}
      \toprule
      \multirow{2}{*}{Setting} & \multirow{2}{*}{Method}
        & \multicolumn{6}{c|}{Self-Reenactment}
        & \multicolumn{2}{c|}{Cross-Reenactment} & \multicolumn{2}{c}{User Study}\\
      \cmidrule(lr){3-12}
        &
        & PSNR $\uparrow$
        & SSIM $\uparrow$
        & LPIPS $\downarrow$
        & FID $\downarrow$
        & FVD $\downarrow$
        & OC $\uparrow$
        & FID $\downarrow$
        & FVD $\downarrow$ 
        & VQ $\uparrow$
        & RF $\uparrow$\\
      \midrule
      \multirow{6}{*}{\shortstack{Short video\\generation}}
       & VACE & 28.60 & 0.937 & 0.056 & 34.83 & 211.2 & 0.880 & 68.67 & 524.7 & 3.942   &  2.796 \\
        & UniAnimate-DiT & 22.20 & 0.754 & 0.179 & 44.01 & 325.1 & 0.846 & 76.40 & 640.5 & 3.496 & 2.971 \\
       & MimicMotion & 20.13 & 0.685 & 0.206 & 48.89 & 395.1 & 0.777 & 68.50 & 608.5 & 2.821 & 2.094 \\
        & HOI-Swap & 24.29 & 0.843 & 0.173 & 50.67 & 352.1 & 0.787 & 76.72 & 570.5 &1.475  &1.201 \\
       & \textbf{Ours} & \textbf{31.71} & \textbf{0.952} & \textbf{0.036} & \textbf{11.53} & \textbf{67.95} & \textbf{0.937} & \textbf{58.83} & \textbf{429.5} & \textbf{4.487} & \textbf{4.636} \\
      \midrule
      \multirow{4}{*}{\shortstack{Long video\\generation}}
       & VACE & 26.32 & 0.937 & 0.054 & 40.83 & 195.9 & 0.882 & 81.77 & 782.0 & 3.143   &  2.287 \\
       & UniAnimate-DiT & 18.94 & 0.695 & 0.295 & 64.08 & 559.7 & 0.823 & 87.46 & 881.7 & 2.061 & 1.506 \\
       & MimicMotion & 17.37 & 0.633 & 0.305 & 59.19 & 488.0 & 0.759 & 85.63 & 776.2 & 2.002 & 1.428 \\
       & \textbf{Ours} & \textbf{30.69} & \textbf{0.951} & \textbf{0.037} & \textbf{9.78} & \textbf{42.17} & \textbf{0.932} & \textbf{57.62} & \textbf{515.0} & \textbf{4.458} & \textbf{4.534}\\
      \bottomrule
    \end{tabular}%
  }
\end{table*}

\subsection{Comparison with other methods}
We compare GenHOI against state-of-the-art methods on both short and long video generation in Tab. \ref{tab:short_video_main_result}. 


\noindent \textbf{Short video generation evaluation} We first evaluate the short video generation capability of each method. As shown, HOI-Swap, the representative HOI reenactment method, fails to generalize to in-the-wild datasets, achieving only 24.29 PSNR in self-reenactment. In contrast, our method generalizes well and achieves state-of-the-art results across all metrics in both self- and cross-reenactment. Compared with the strongest competitor VACE, our method outperforms it on all metrics by a clear margin. This improvement mainly stems from our superior ability to maintain object consistency. User study results in Tab.~\ref{tab:short_video_main_result} further confirm this claim, showing a clear advantage over VACE in reference fidelity (4.6 vs. 2.79).

We visualize cross-reenactment results in Fig. \ref{fig:swap} (down). The proposed method successfully handles objects of various shapes and sizes, enabling transformations between large and small objects as well as different object types. For example, transforming a bag into a lipstick, a milk tea cup, or even a non-existent magical wand. Such flexibility is highly useful in practical applications, significantly reducing production costs in e-commerce scenarios. Fig. \ref{fig:swap} (up) presents a visual comparison with existing methods. As illustrated, previous methods often fail to maintain temporal or object consistency, whereas our approach generates natural and visually coherent human–object interactions across videos.

\noindent \textbf{Long video generation evaluation} The proposed method achieves an even larger performance lead in long video generation. Compared with the second-best method, VACE, the PSNR gap in self-reenactment increases from 3.11 to 4.37, the FVD gap in cross-reenactment expands from 95.2 to 267, and the user study video quality gap rises from 0.5 to 1.3. These results further confirm the superiority of our method in producing temporally coherent and high-quality long videos.

\begin{table}[!t]
  \centering
  \caption{Ablation study on the components of proposed method. HS RoPE denotes head-sliding RoPE, SAG is spatial attention gate and FlF denotes first-last-frame condition.}
   \vspace{-0.1in}
  \label{tab:ablation_2}
  \small
  \setlength{\tabcolsep}{2.5pt}
  \resizebox{0.47\textwidth}{!}{%
    \begin{tabular}{l|cccccc}
      \toprule
      Methods 
        & PSNR $\uparrow$
        & SSIM $\uparrow$
        & LPIPS $\downarrow$
        & FID $\downarrow$
        & FVD $\downarrow$
        & OC $\uparrow$ \\
      \midrule
      HCU   
        & 28.25 & 0.942 & 0.058 & 22.89 & 248.6 & 0.907\\
      HCU + separate RoPE      
        & 29.73 & 0.945 & 0.050 & 22.66 & 223.8 & 0.908\\
      HCU + ref-in-bbox    
        & 30.34 & 0.945 & 0.044 & 18.23 & 101.9 & 0.919\\
      HCU + HS RoPE    
        & 30.88 & 0.951 & 0.039 & 17.92 & 103.9 & 0.915\\ 
      HCU + HS RoPE + SAG 
        & 31.21 & 0.952 & 0.038 & 16.79 & 98.09 & 0.920\\
      HCU + HS RoPE + SAG + FLF(full)
        & \textbf{31.71} & \textbf{0.952} & \textbf{0.036} & \textbf{11.53} & \textbf{67.95} & \textbf{0.937}\\
      \bottomrule
    \end{tabular}%
  }
\end{table}

\subsection{Ablation Study}

We perform various ablation studies to show the effectiveness of each component in our network. 
All experiments are reported on the AnchorCrafter\_HOI dataset in self-reenactment setting. HCU denotes we finetune the video generation model with the proposed HOI condition unit, which is considerate as our baseline. Results are shown in Table \ref{tab:ablation_2}. Below we describe each component in more detail.


\vspace{0.05in} 


\vspace{0.05in}
\noindent \textbf{Head-Sliding RoPE vs. ref-in-bbox vs. 3D RoPE} As the strategy for injecting reference-object information is central to our approach, we provide a detailed comparison of three alternatives. “Separate RoPE” denotes the separate rotary positional embedding used in Stand In \cite{xue2025stand} and OminiControl \cite{ominicontrol}, while ref-in-bbox \cite{rehold} follows the formulation in Eq.~\ref{eq:ref_in_bbox}. As shown by our results, the proposed Head-Sliding RoPE improves PSNR by 1.15~dB over separate RoPE and by 0.54~dB over ref-in-bbox. Similar gains are observed across the other evaluation metrics. We attribute these improvements to Head-Sliding RoPE’s ability to effectively average the attention responses of conditional tokens over the full temporal span of the video. 

\noindent \textbf{Impact of the Spatial Attention Gate.} Adding the spatial attention gate increases PSNR from 30.88 to 31.12~dB and reduces FVD from 103.9 to 98.09 (lower is better), underscoring the necessity of imposing a spatial constraint.

\noindent \textbf{Impact of the First–Last-Frame Condition.} The first–last-frame (FLF) conditioning strategy is commonly used in video generation models \cite{wan2025wanopenadvancedlargescale}. In our framework, enabling FLF raises PSNR from 31.21 to 31.71 dB, indicating that anchoring the sequence with boundary frames benefits generation quality.

\noindent \textbf{Model Parameter Analysis.} 
Overall, the proposed module augments the pretrained video generation model with only 157M additional learnable parameters, accounting for approximately 
0.95\% of the original 16.5B-parameter model. This lightweight design offers two main advantages: (1) it largely preserves the original model’s generalization ability, and (2) it introduces minimal computational overhead. Consequently, our model can achieve state-of-the-art HOI reenactment performance using only limited training data.



\begin{figure*}[h]
  \centering
  \includegraphics[width=0.9\linewidth]{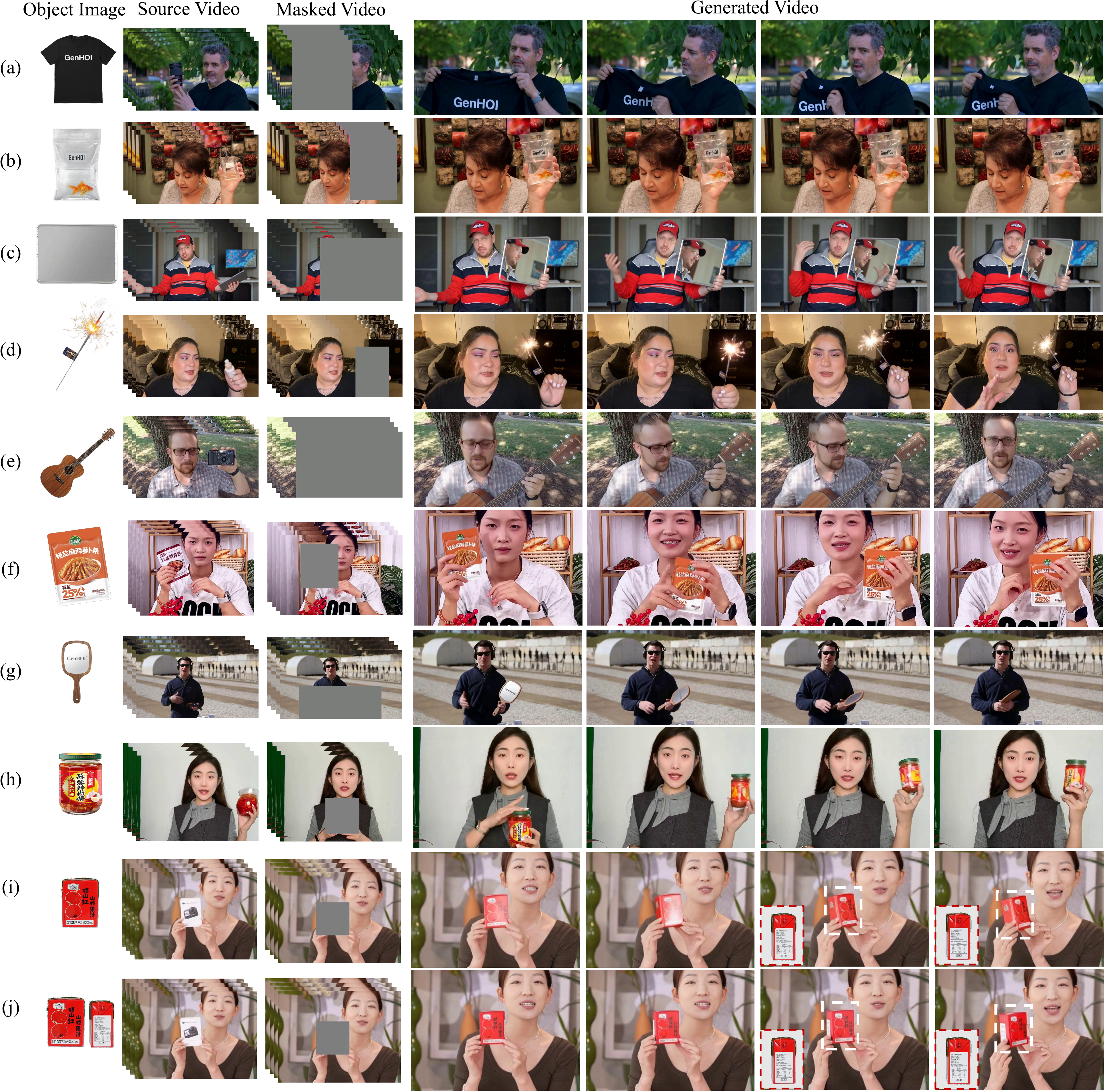}
  \caption{Additional qualitative results including deformable objects (a,b), dynamic physics (b,c,d), complex interactions (e,f), robustness to occlusions during generation (f), and object rotation (g-j). Better viewed by zooming in.}
  \label{fig:rebuttal}
\end{figure*}

\subsection{More Results and Analysis}



\begin{figure*}[h]
  \centering
  \includegraphics[width=1.0\linewidth]{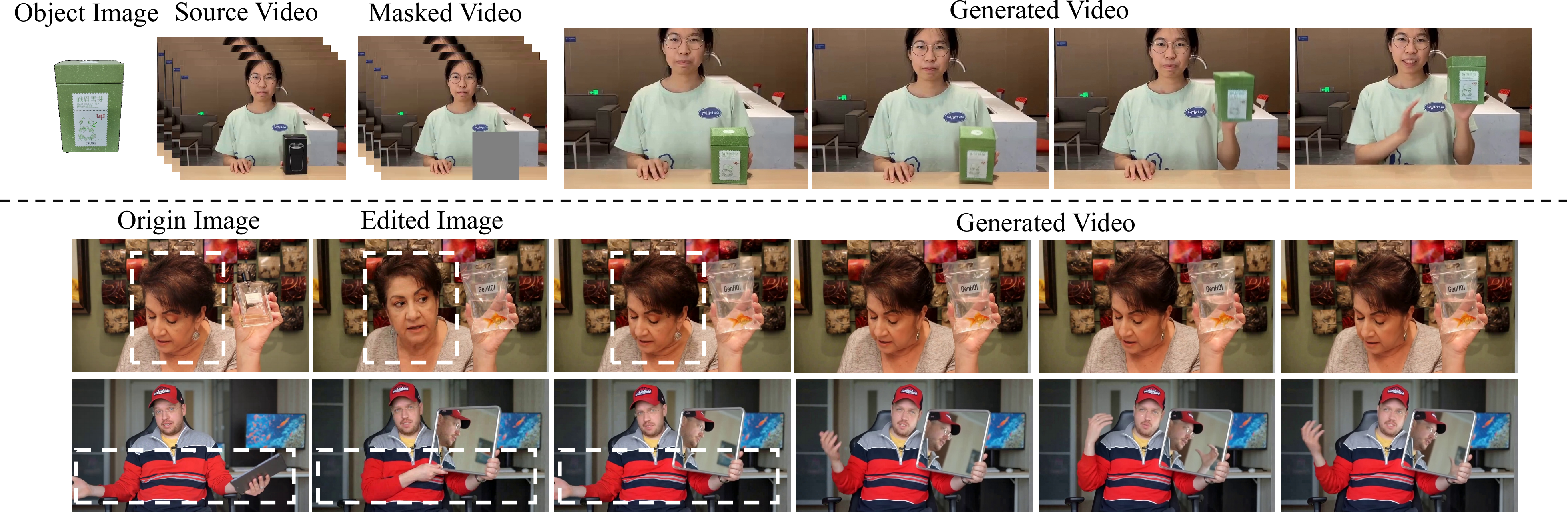}
  \caption{Top: Cross-reenactment results without an edited first frame, showing that our method remains robust and preserves object identity using only the reference image. Bottom: Robustness to first-frame edits—despite pose or position shifts in the edited first frame, the generated videos maintain consistent object appearance and interaction.}
  \label{fig:first_frame}
\end{figure*}



\begin{figure*}[h]
  \centering
  \includegraphics[width=1.0\linewidth]{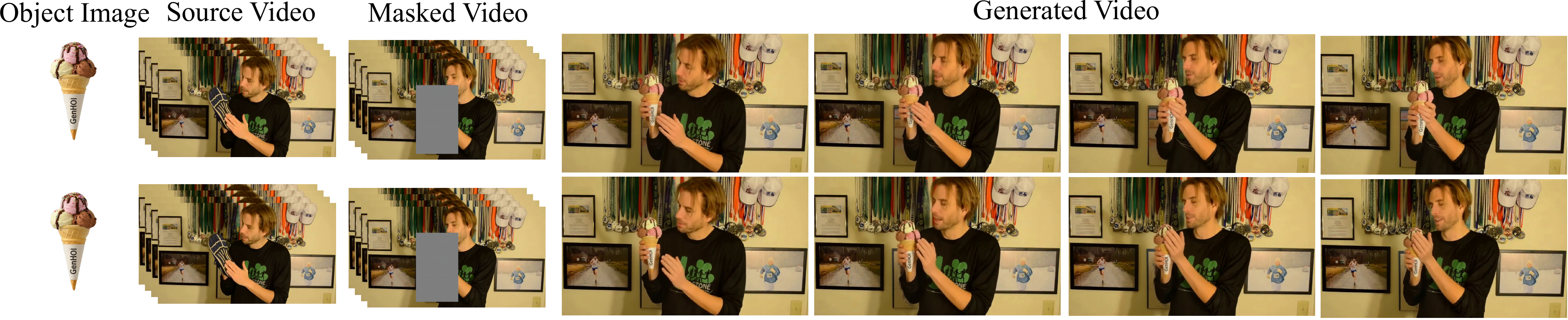}
  \caption{Diversity across repeated sampling with the same inputs, illustrating variations in motion and contact patterns produced without explicit pose constraints.}
  \label{fig:diversity}
\end{figure*}

\noindent \textbf{Challenging Cases}
Fig.~\ref{fig:rebuttal} presents additional manipulation results beyond the AnchorCrafter evaluation dataset, including deformable objects (Fig.~\ref{fig:rebuttal} (a,b)), dynamic physics (Fig.~\ref{fig:rebuttal} (b,c,d)), complex interactions (Fig.~\ref{fig:rebuttal} (e,f)), robustness to occlusions during generation (Fig.~\ref{fig:rebuttal} (f)), and object rotation (Fig.~\ref{fig:rebuttal} (g-j)). These cases highlight that our model maintains stable hand--object contact and preserves object identity under large deformation and dynamic motion. The results also indicate robustness to partial occlusions, where the object remains coherent across frames. It also handles viewpoint changes during rotation well, preserving the overall geometry even for unseen views.

\noindent \textbf{Multi-view Condition \& rotation}
The proposed method can generate unseen object views using single-image condition, making it more user friendly. Fig.~\ref{fig:rebuttal} (g,h) demonstrates its behavior under rotation, where the results are plausible and preserve object category and geometry with single object image. However, generating unseen views remains dependent on the model's generative priors, which may lead to texture inconsistencies with the target object. For example, in Fig.~\ref{fig:rebuttal} (i), the synthesized textures in unseen views (white dashed boxes) differ from those of the original object (red dashed boxes). This limitation can be alleviated by introducing multi-reference conditioning, where multi-view inputs significantly improve texture consistency (Fig.~\ref{fig:rebuttal} (j)).

\noindent \textbf{Influence of first-frame editing quality.} The edited first frame mainly serves as auxiliary information, providing a scale cue for the object. This is because our inpainting-based design and the proposed Head-Sliding mechanism extract texture features directly from the object reference image rather than the edited first frame. Consequently, our method can generate reasonable videos \textbf{even without an edited first frame}, as shown in Fig.~\ref{fig:first_frame} (top). For the same reason, the final videos are robust to pose or position shifts introduced by the initial edit, as shown in Fig.~\ref{fig:first_frame} (bottom).

\noindent \textbf{Mask boundary influence.} Our method is also robust to variations in mask precision. As shown in Fig.\ref{fig:rebuttal}(d,e), both precise and coarse masks produce high-quality results. Larger masks generally induce greater changes from the original video, but the mask should at least cover the original hand–object interaction region.

\subsection{Limitation}
Although GenHOI demonstrates strong performance in HOI scenarios, it still exhibits certain limitations that call for further study and improvement.
\begin{itemize}
\item In the first–last frame mode, the quality of the generated video largely depends on the physical plausibility of the given first and last frames. Existing image editing models struggle to guarantee this, especially when the hand pose in the first frame needs to transition naturally to the pose in the last frame within a very short time.

\item For challenging manipulations or non-rigid objects, the output quality may vary across different runs, reflecting the inherent ambiguity of these scenarios. How to consistently produce satisfactory results in such complex cases remains an open problem.
\end{itemize}

\section{Conclusion}
In this work, we present GenHOI, a novel framework designed to advance realistic hand–object interaction reenactment. Specifically, we introduce a HOI condition unit that adapts existing video generation models to the HOI reenactment task. Furthermore, we propose head-sliding RoPE and a spatial attention gate to achieve temporally balanced and spatially selective object information injection. Together, these mechanisms enable the model to maintain natural hand–object interactions and preserve object identity, even in unseen, in-the-wild scenarios. Comprehensive experiments demonstrate that GenHOI consistently outperforms existing HOI reenactment and all-in-one video editing methods in both qualitative and quantitative evaluations.

{
    \small
    \bibliographystyle{ieeenat_fullname}
    \bibliography{main}
}


\end{document}